\documentclass[letterpaper, 10 pt, journal, twoside]{ieeetran}

\IEEEoverridecommandlockouts                              % This command is only needed if 
\usepackage{booktabs} 
\usepackage{multirow}
\usepackage{amsmath} % assumes amsmath package installed
\usepackage{amssymb}
\usepackage{graphicx}
\usepackage{dblfloatfix} % fixes the placement of table* and figure*
\usepackage{pifont} % checkmarks
\usepackage{svg}
\usepackage{subcaption}
\usepackage{makecell}
\newcommand{\cmark}{\ding{51}}%
\newcommand{\xmark}{\ding{55}}%

\usepackage[noadjust]{cite}

\usepackage{xcolor}

\title{\LARGE \bf
PAg-NeRF: Towards fast and efficient end-to-end panoptic \\ 3D representations for agricultural robotics}

\author{Claus Smitt$^{\dag}$, Michael Halstead$^{\dag}$, Patrick Zimmer$^{\dag}$, Thomas Läbe$^{\ddag}$, Esra Guclu$^{\dag}$, \\ Cyrill Stachniss$^{\ddag}$,  Chris McCool$^{\dag}$% <-this % stops a space
\thanks{Authors \dag\ are with Institute of Agriculture, Agriculture Robotics \& engineering and \ddag\ are with the Institute of Photogrametry of the University of Bonn, Germany
{\tt\small [csmitt, patrick.zimmer, michael.halstead, egueclue, cmccool]@uni-bonn.de, laebe@ipb.uni-bonn.de, cyrill.stachniss@igg.uni-bonn.de}}%
}

\begin{document}

\maketitle
\thispagestyle{empty}
\pagestyle{empty}

%%%%%%%%%%%%%%%%%%%%%%%%%%%%%%%%%%%%%%%%%%%%%%%%%%%%%%%%%%%%%%%%%%%%%%%%%%%%%%%%
\begin{abstract}
Precise scene understanding is key for most robot monitoring and intervention tasks in agriculture. 
In this work we present \textbf{PAg-NeRF} which is a novel NeRF-based system that enables 3D panoptic scene understanding.
Our representation is trained using an image sequence with noisy robot odometry poses and automatic panoptic predictions with inconsistent IDs between frames.
Despite this noisy input, our system is able to output scene geometry, photo-realistic renders and 3D consistent panoptic representations with consistent instance IDs.
%Yet, our model is able to estimate a panoptic 3D representation with consistent instance IDs.
We evaluate this novel system in a very challenging horticultural scenario and in doing so demonstrate an end-to-end trainable system that can make use of noisy robot poses rather than precise poses that have to be pre-calculated. 
%The outputs of this system are scene geometry, photo-realistic renders and 3D consistent panoptic representations.
Compared to a baseline approach the peak signal to noise ratio is improved from 21.34dB to 23.37dB while the panoptic quality improves from 56.65\% to 70.08\%.
Furthermore, our approach is faster and can be tuned to improve inference time by more than a factor of 2 while being memory efficient with approximately 12 times fewer parameters.

%\comment{You almost always finish the abstract with the key performance numbers.}
%\updated{This approach is able to accurately estimate the number of fruit (sweet pepper) in a glasshouse with a normalised absolute error of 5.1\% and an $R^{2}$ of 0.902 with the visual ground truth.}

\end{abstract}

%%%%%%%%%%%%%%%%%%%%%%%%%%%%%%%%%%%%%%%%%%%%%%%%%%%%%%%%%%%%%%%%%%%%%%%%%%%%%%%%

\section{Introduction}

% \todo{An example of leaving a todo in the paper}
% \comment{An example of leaving a comment in the paper}
% \updated{An example of highlighting where the text has been updated (use this after the 1st draft is done).}

In recent years the agricultural sector has rapidly incorporated multiple robotic systems to perform monitoring and intervention tasks~\cite{ahmadi2021autonomous, you2022semantics, Lehnert20_1, crop_agnostic_halstead2021crop, sa2016deepfruits, halstead2018fruit, zabawa2019detection}.
This is due to emerging needs of a more efficient and sustainable production, driven by factors such as climate change, scarcity of skilled labour, customer requirements, and increasing production costs.
%~\cite{somethingPlantSciency}.
The successful adoption of robotic systems in agriculture is largely due to recent advancements in vision based deep learning (DL)~\cite{sa2016deepfruits,halstead2018fruit, smitt2022explicitly}.
In particular, the ability to perform vision based semantic and spatial reasoning in the robot's environment. % to automate agricultural tasks.

\begin{figure}[th!]
\centering
    \includegraphics[width=.36\textwidth]{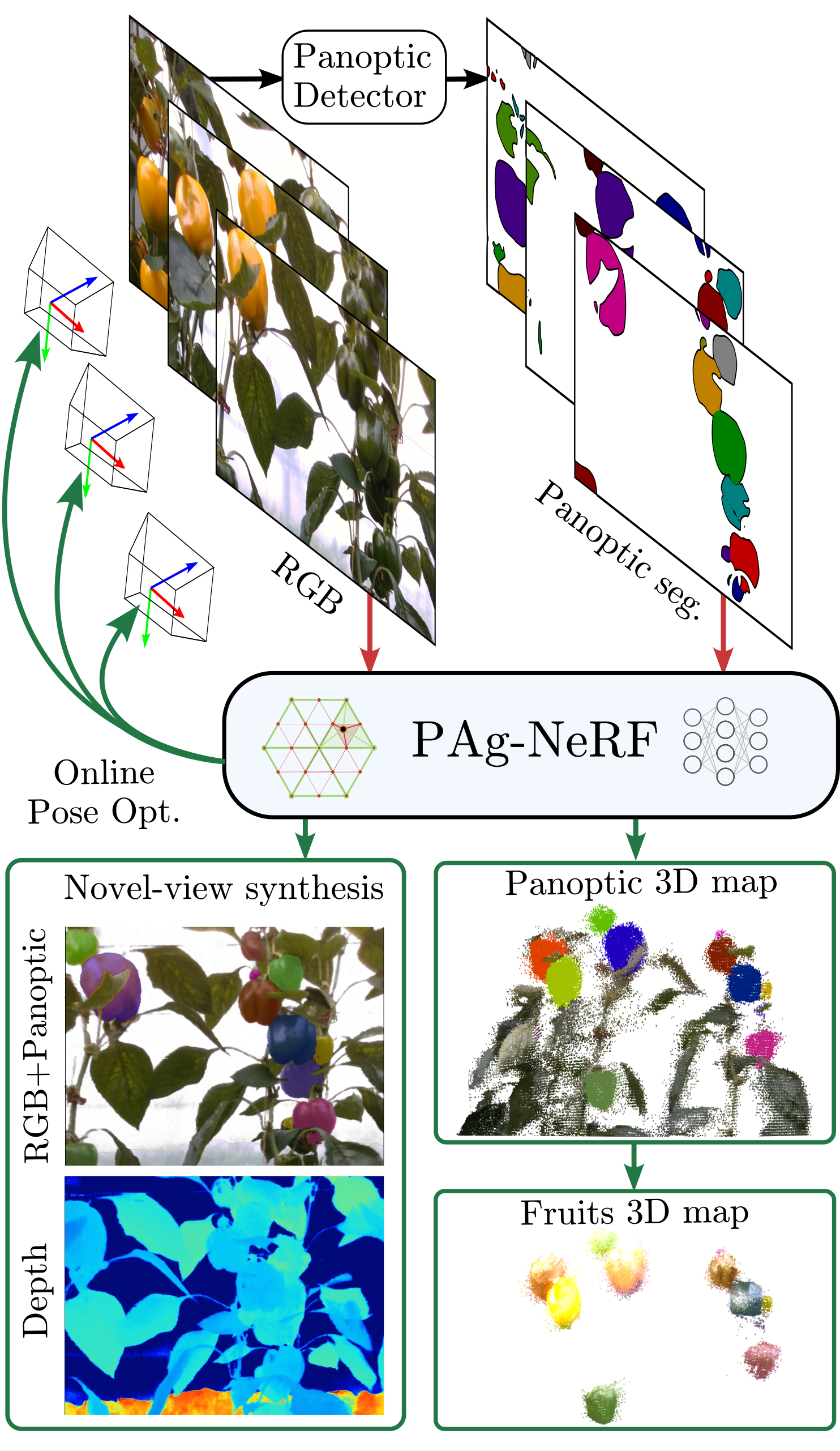}
    \caption{PAg-NeRF is a fast and efficient model that renders novel-view photo-realistic images and ID consistent panoptic 3D maps from images, panoptic detections and noisy poses.}
    \vspace{-20pt}
\label{fig:hero}
\end{figure}

In horticulture, detecting~\cite{sa2016deepfruits}, measuring size~\cite{crop_agnostic_halstead2021crop}, estimating ripeness~\cite{halstead2018fruit} and counting fruit~\cite{smitt2021pathobot} are some key monitoring tasks that provide growers detailed information to make better decisions, improving sustainability, and increasing production efficiency.
Current state-of-the-art vision systems for detection and ripeness estimation use DL to perform instance-based semantic segmentation~\cite{crop_agnostic_halstead2021crop}.
%Currently the detection and ripeness estimation are tackled with state-of-the-art vision systems using DL to perform instance-based semantic segmentation~\cite{crop_agnostic_halstead2021crop}.
%Moreover, incorporating 3D information can greatly improve fruit counting and size estimation~\cite{smitt2022explicitly,crop_agnostic_halstead2021crop,smitt2021pathobot,yuepan2023panoptic}.
A set of recent work show that incorporating 3D information can greatly improve fruit counting and size estimation~\cite{smitt2022explicitly,crop_agnostic_halstead2021crop,smitt2021pathobot,yuepan2023panoptic}.
%Nonetheless, horticultural environments such as sweet pepper cropping remain extremely challenging environments due to the high level of occlusion, illumination changes, and green over green detection scenarios, among others.
Despite these advances horticultural environments (e.g. sweet pepper cropping) remain extremely challenging due to the high level of occlusion, illumination changes, and green over green detection scenarios.

Recently, neural radiance fields (NeRF) have shown great potential to implicitly represent 3D information from just posed images.
%Since their inception, neural radiance fields (NeRF) have become a very popular approach to obtain implicit 3D scene representation from posed images only.
%This representation has been used to predict multiple properties of 3D scenes such as geometry~\cite{rosu2023permutosdf}, photo-realistic appearance~\cite{barron2021mip}, and more recently semantics~\cite{semantic_nerf_zhi2021place,zhi2021ilabel}.
NeRF has been used to predict multiple properties of 3D scenes such as geometry~\cite{rosu2023permutosdf}, photo-realistic appearance~\cite{nerf_mildenhall2021nerf}, and more recently semantics~\cite{semantic_nerf_zhi2021place} that are consistent across novel views.
Furthermore, recent contributions have considerably improved their performance and memory efficiency~\cite{neuralSVF_liu2020neural,rosu2023permutosdf,muller2022instant}, turning them into a promising representation for robotics applications.

In this work, we present a novel system \textbf{PAg-NeRF} (Fig.\ref{fig:hero}) which performs online pose optimization~\cite{lin2021barf} using state-of-the-art accelerated NeRFs~\cite{rosu2023permutosdf} to produce 3D consistent panoptic representations.
These representations enable us to resolve identity associations across image sequences using only the implicit geometry of NeRF.
We deploy this novel system in a very challenging horticultural scenario and in doing so make the following contributions.

%\begin{enumerate}
First, \textbf{we propose an end-to-end trainable system} that makes use of noisy robot poses rather than precise poses that have to be pre-calculated. 
The outputs of this system are scene geometry, photo-realistic appearances, and 3D consistent panoptic representations.

Second, \textbf{we demonstrate this system is a fast and efficient 3D panoptic radiance architecture}.
It is validated on real-world agricultural data and outperforms previous work~\cite{panoptic_lifting_siddiqui2023panoptic}, while being twice as fast at inference with approximately 12 times fewer parameters.

%\end{enumerate}

%3D panoptic estimation (consistent indentities across frames)~\cite{panoptic_lifting_siddiqui2023panoptic} to produce 3D consistent panoptic representations in a very challenging horticultural scenario.
%which is a fast and memory efficient architecture that leverages state-of-the-art accelerated NeRFs~\cite{rosu2023permutosdf}, online pose optimization~\cite{lin2021barf}, and panoptic baking approaches~\cite{panoptic_lifting_siddiqui2023panoptic} to produce 3D consistent panoptic representations in a very challenging horticultural scenario.
%In this work we present \textbf{PAg-NeRF} which is a fast and memory efficient architecture that leverages state-of-the-art accelerated NeRFs~\cite{rosu2023permutosdf}, online pose optimization~\cite{lin2021barf}, and panoptic baking approaches~\cite{panoptic_lifting_siddiqui2023panoptic} to produce 3D consistent panoptic representations in a very challenging horticultural scenario.
%Moreover, our model tackles issues typically encountered in agricultural environments for a successful reconstruction of the scene.
%The main contribution of our work is that:
%This leads to our work's main contributions being:
%\begin{itemize}
%    \item A fast and efficient 3D panoptic radiance architecture;
%    \item End-to-end trainable, even with noisy robot poses;
%    \item Validated on a challenging horticulture scenario;
%    \item As well as speed and memory ablation studies.
%\end{itemize}

\section{Related Work}
A diverse range of techniques have been employed in agriculture to provide robotic systems with geometric and semantic scene understanding.
More recently, NeRF approaches have also been used to enhance monitoring systems in the agriculture sector.
Below we provide a brief review of the relevant monitoring systems and recent advancements in fast and efficient NeRFs that allowed the development of this work.

\subsection{Crop Monitoring}
Crop monitoring is an important component of any agricultural robotic platform, from arable farmland to glasshouses.
Without accurate monitoring, these platforms would be unable to accurately perform intervention activities such as weeding~\cite{ahmadi2021autonomous}, harvesting~\cite{Lehnert20_1}, or yield estimation~\cite{crop_agnostic_halstead2021crop}.
% Performing these tasks with robots decreases the requirements on humans and has the potential to increase farming revenue by streamlining these processes and decreasing costs assigned to manual labor.

In recent years DL, in particular deep neural networks (DNNs), has dominated state-of-the-art techniques for agricultural monitoring. % leading to large steps in applicability.
% For deep neural networks (DNNs), the majority of early approaches to agricultural monitoring concentrated on top-down approaches such as Faster-RCNN~\cite{ren2015faster}.
Sa \textit{et al.}~\cite{sa2016deepfruits} showed how Faster-RCNN~\cite{ren2015faster} could be fine-tuned for accurate sweet pepper detection.
This was extended in~\cite{halstead2018fruit} to include subclass-based (fruit ripeness) classification, integrated into a fruit tracking approach to count fruit. % produce a fruit count. % estimation in a field.
%In the work of Halstead \textit{et al.}~\cite{halstead2018fruit} they also track fruit through the scene to produce a yield estimate.
%Using a complex multi-task cascaded convolutional network Zhang \textit{et al.}~\cite{zhang2019multi} was able to detect fruit in an orchard accurately, however, this was a complex network architecture.
Turning grape detection into a three-class problem (background, grapes, edges) Zabawa \textit{et al.}~\cite{zabawa2019detection} improved the detection of grapes in an orchard by creating better distinction between the individual grapes. 
%As an investigation into cross-domain performance~\cite{halstead2020} showed that Mask-RCNN~\cite{he2017mask} was able to improve sweet pepper classification and detection accuracy when compared to Faster-RCNN.
Once again using a top-down approach to instance-based semantic segmentation, ~\cite{crop_agnostic_halstead2021crop} showed that crop monitoring could use a similar approach in both arable farmland and glasshouses.
% In a data-driven manner they were able to accurately segment either fruit or plants in a scene and relay basic phenotyping (class and area) information to the end user; their approach also improved yield estimation in both scenarios by introducing a more dynamic method for multiple object tracking.
In general, the majority of these approaches are still-image based and do not consider spatial-temporal information.
Smitt \textit{et al.}~\cite{smitt2022explicitly} showed how integrating robot trajectory and 3D scene information into DNNs could improve state-of-the-art results. %can improve semantic segmentation, improving state-of-the-art results.
More recently~\cite{yuepan2023panoptic} used instance detections of fruit which were tracked and mapped into a multi-resolution occupancy grid.
Fruit 3D models were also predicted with a CNN and aligned with the grid, generating a panoptic 3D map of the crops.
% On very recent work~\cite{yuepan2023panoptic} instance detections of fruits were tracked and mapped into a multi-resolution occupancy grid, and fruit 3D models were predicted with a CNN and aligned with the grid, generating a panoptic 3D map of the crops.

% \todo{These approaches all use top-down segmentation... link better to the next subseection.}

\subsection{Panoptic Segmentation}

\begin{figure}[!t]
    \centering
    \includegraphics[width=1\linewidth]{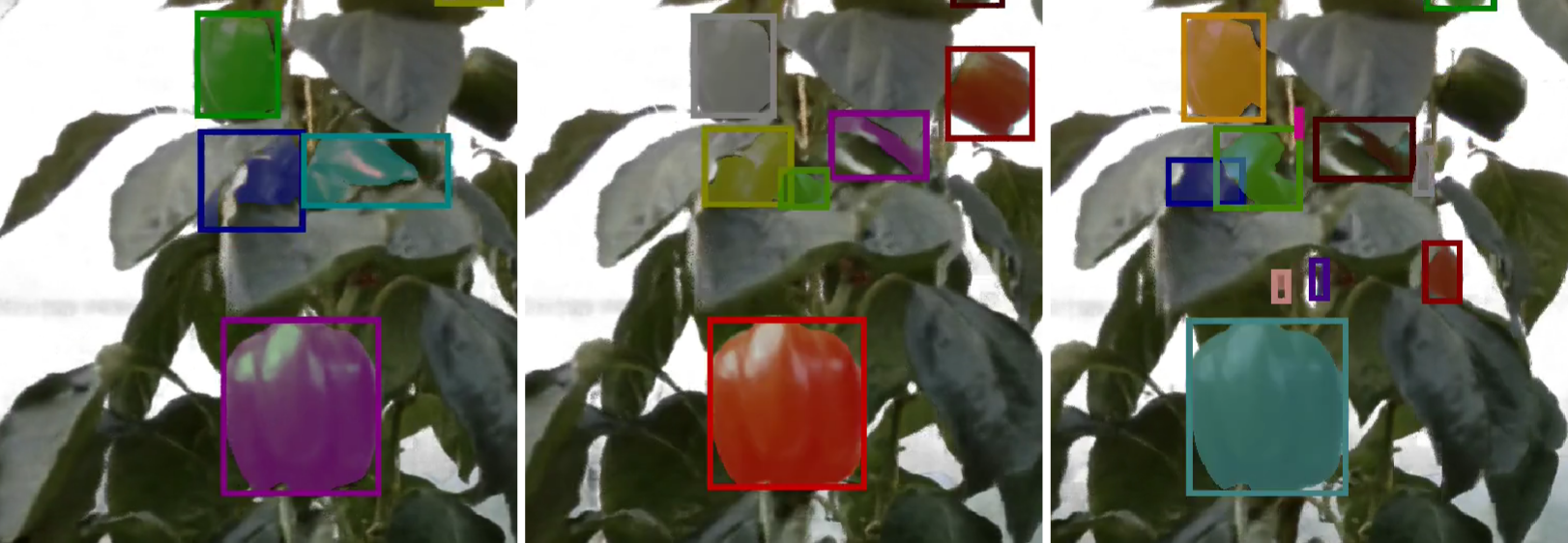}
    \caption{Still-image panoptic detector producing inconsistent instance IDs and false positives in a sequence of frames}
    \label{fig:panopticFlicker}
    \vspace{-17pt}
\end{figure}

Panoptic segmentation jointly solves the tasks of semantic and instance segmentation and combines them in a single prediction.
A semantic label is predicted for each pixel of ``stuff'' segments, as well as an instance ID for ``things'' segments, yielding detailed and effective scene understanding~\cite{panopticSegMatric_kirillov2019}.
In early works, semantic and instance outputs of CNNs were combined to obtain panoptic predictions~\cite{panopticFPN_kirillov2019,panopticdeeplab_cheng2020}.

% adds a dense prediction branch to the Mask-RCNN~\cite{he2017mask} architecture to perform panoptic segmentation. In this way, instance and semantic segmentation tasks are performed separately and the outputs are fused. Cheng \textit{et al.}~\cite{panopticdeeplab_cheng2020} propose Panoptic-DeepLab that utilizes object center predictions with a clustering algorithm to perform instance segmentation. Then semantic and instance predictions are combined. Xiong \textit{et al.}~\cite{upsnet_xiong2019} introduce UPSNet which employs a parameter-free panoptic head at the end of the network to get more accurate results. K-Net of Zhang \textit{et al.}~\cite{kNet_zhang2021} uses dynamic kernels to predict panoptic masks.

% , and Wang \textit{et al.}~\cite{maxdeeplab_wang2021} developed MaX-DeepLab by combining CNN and transformer modules together to produce panoptic predictions.
% They extend Axial-DeepLab~\cite{axialdeeplab_wang2020} to predict instance masks directly.
More recently, vision-based transformers have gained popularity with their strong performance~\cite{transformersurvey_khan2022} such as Mask2Former~\cite{mask2former_cheng2022masked}.
This model was proposed as a transformer-based universal segmentation model, which uses masked-attention to extract localized features.
Another recent method by Jain \textit{et al.}~\cite{oneformer_jain2023} applies task-guided queries to obtain mask predictions, achieving state-of-the-art performance.
% \todo{MAH: this feels a little weak. I'd add some information. I'd also link it better to the next paragraph.}.

These techniques are an obvious option for crop monitoring purposes.
However, they are still-image detectors and do not produce consistent instance ID predictions between frames (as depicted in Fig.~\ref{fig:panopticFlicker}).
The flickering instance ids can be tackled separately by solving instance tracking~\cite{crop_agnostic_halstead2021crop, yuepan2023panoptic, smitt2021pathobot}.
By contrast, in this work we use frame-based instance predictions to train a NeRF 3D scene representation which then implicitly ensures there is inter-frame ID consistency.

%directly fuse the frame-based instance predictions into a NeRF 3D scene representation, implicitly solving inter-frame ID consistency.

\subsection{Neural Randiance Fields}
Since their original introduction~\cite{nerf_mildenhall2021nerf}, NeRFs have become a very popular method to implicitly represent 3D scenes from posed views.
Some applications include photo-realistic novel-view synthesis~\cite{nerf_mildenhall2021nerf}, localization and mapping~\cite{lin2021barf,sucar2021imap}, detailed 3D reconstruction~\cite{rosu2023permutosdf} and semantic 3D mapping among others~\cite{dellaert_kundu2022panoptic, panoptic_lifting_siddiqui2023panoptic,geiger_fu2022panoptic}.
The original NeRF implementation leveraged differentiable rendering~\cite{volumetric_niemeyer2020differentiable} to represent 3D scenes using fixed positional encodings (PEs) to encode positions and view directions and decode them with deep multi layer perceptrons (MLPs)~\cite{nerf_mildenhall2021nerf}.
Such methods can tackle several tasks, but are very memory and compute intensive, and use offline structure from motion (SFM) to compute camera poses.

% As shown in \cite{nerf}, positional encodings are key for learning high detail radiance fields.
% This concept was extending by employing learnable multi-scale grid encodings which are suitable for representing spatial information \cite{instantNGP,permutogrids,pointcloudNerfIPB}.
% Moreover, by interpolating spatial-queried points only and propagating gradients only them, such encodings allow for faster training and inference.
% To further increase running time performance, occupancy grids can be employed \cite{firstNerfOcuppancy,InstanceNGP,permuthoGrids}, where only voxels with high density are considered occupied.
% This way, during the ray-tracing process, only ray samples falling in occupied voxels are used to query the model.  

%\subsubsection{Learnable geometric encodings}
\subsubsection{Learning efficient geometric encodings}
Seeking to improve training and inference speed of NeRFs~\cite{neuralSVF_liu2020neural} bookkeep an occupancy grid to sample only occupied space, maintaining however a large MLP network.
More recently Sun et al.~\cite{dvgo_sun2022direct} employed 3D grids with learnable features at their vertices and a very shallow MLP decoder on top.
This allowed the gradients to only be propagated through the interpolated features when queried by ray-tracing.
In~\cite{chen2022tensorf} 3D grids were decomposed into planes and vectors, reducing drastically the number of learnable parameters, without sacrificing performance.
Takikawa \textit{et al.}~\cite{lod_takikawa2021neural} use multi-resolution grids to efficiently represent different levels of detail (LOD) of the scene without the need for positional encoding.
M{\"u}ller \textit{et al.}~\cite{muller2022instant} embedded sparse hash tables in multi-resolution grids, with less parameters than keys.
Hash collisions were resolved through the training process, yielding implicitly sparse grids to learn fast and memory efficient neural graphic models.

More recently, Rosu \textit{et al.}~\cite{rosu2023permutosdf} tackle fast estimation of high-detailed sign distance functions (SDF).
They employed permutoedral hash-grids, instead of cubic ones, since they interpolate less points per queried 3D coordinate resulting in faster training and inference.
This method also leverages the implicit sparseness of under-parameterized hash tables making them memory efficient as well.

\subsubsection{3D semantic NeRFs}
The continuous differentiable nature of NeRFs allows encoding of any continuous properties into 3D space.
Recently Zhi \textit{et al.}~\cite{semantic_nerf_zhi2021place} presented an approach to encode semantics into an MLP based NeRF with noisy 2D predictions of synthetic environments as input.
This yielded cleaner and 3D consistent novel-view semantic predictions.
% Another approach~\cite{zhi2021ilabel} baked DNN features in a NeRF representations.
% By using sparse user click labels they train shallow semantic decoders that predict the annotated classes in the whole scene, due to feature similarities.

In the field of horticulture monitoring, Kelly \textit{et al}~\cite{kelly2023target} trained a MLP NeRF model with strawberries and sweet pepper images captured by field robots.
Semantic detections were used to refine the render quality of fruits, and inspired by~\cite{lin2021barf} they jointly optimize camera poses, using odometry as initial guess, avoiding the need for offline SFM.
Later works encoded instance prediction from 2D images into NeRFs in the context of autonomous cars~\cite{geiger_fu2022panoptic, dellaert_kundu2022panoptic}, handling inter-frame ID inconsistencies with a separate 3D tracking phase.
Very recently, a tracking free panoptic NeRF model was introduced~\cite{panoptic_lifting_siddiqui2023panoptic}, which matches instance ID detections and model outputs by solving a linear assignment problem optimally.
This model uses a TensoRF~\cite{chen2022tensorf} feature encoding grid and is evaluated in indoor environments.

All these panoptic approaches require off-line pose optimization to produce accurate results.
To the best knowledge of the authors, there are no panoptic NeRF end-to-end models that jointly optimize poses and the 3D representation.

\section{Proposed Approach}

%\comment{CM to CS: I think that all of the equations take a lot of space, we should talk about how to cut some of the out.}

Our method generates implicit 3D representations of static environments with color, depth and panoptic segmentation modalities.
We achieve this by training a NeRF model from color frames with automatic (per-frame) panoptic detections and noisy robot poses.
In addition to a regular color grid, we propose a novel delta grid for panoptic decoding.
The resultant model can then be used to perform novel-view synthesis with consistent label IDs for the instances present in a scene.
We target our approach to challenging agricultural scenarios.
~
\begin{figure*}[!ht]
    \vspace{12pt}
    \centering
    \includegraphics[width=0.9\textwidth]{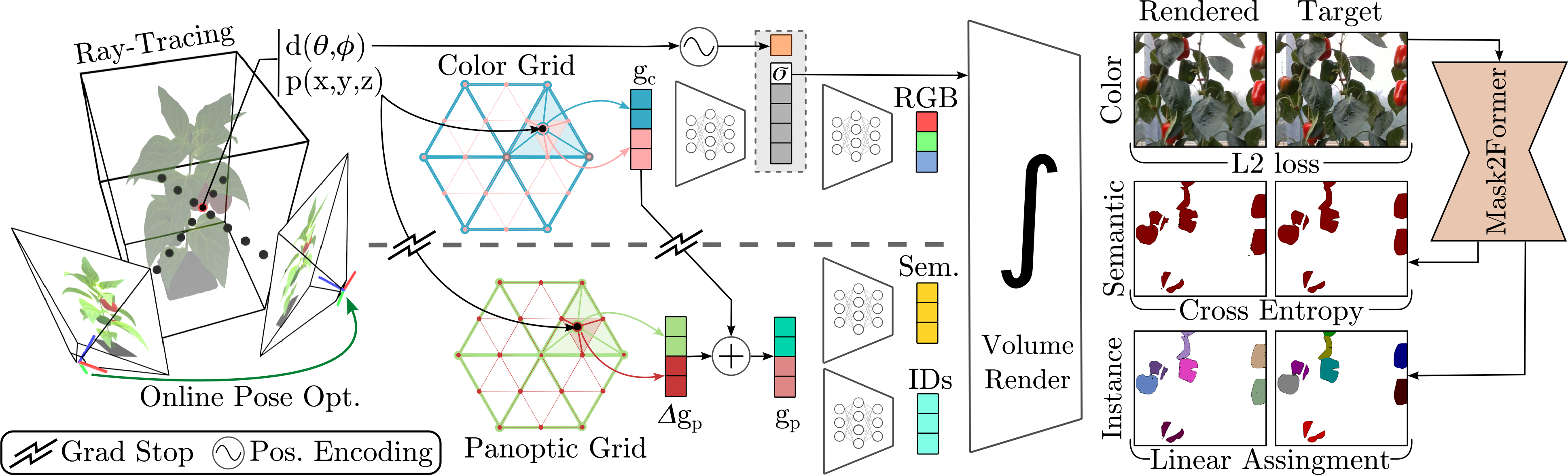}
    \caption{Our 3D neural representation uses 2 separate grids to represent color and panoptic quantities.
    It learns the panoptic representation from automatic detections and the camera poses are jointly optimized with only the color loss.}    
    % \caption{Our 3D neural representation uses 2 separate grids to represent appearance and panoptic quantities, and learns the panoptic representation from detections from the corresponding color images. Camera poses are jointly optimized with the representation according to the color loss only.}
    \label{fig:pipeline}
    \vspace{-16pt}
\end{figure*}
~

%\subsection{Panoptic neural and volumetric rendering}
We represent 3D scenes as volumetric panoptic radiance fields.
These map 3D points $\mathbf{p} \in \mathbb{R}^{3}$ and view directions $\mathbf{d} \in \mathit{S^{2}}$ to volumetric fields with density $\hat{\sigma} \in \mathbb{R}_{[0,\infty]}$, color $\hat{\mathbf{c}} \in \mathbb{R}^{3}_{[0,1]}$ and distributions of $\hat{\mathbf{s}}$ over $\mathit{D}$ semantic classes as well as unique instance ID of objects $\hat{\mathbf{k}}$ over $\mathit{N}$ instances.
%Here $\mathit{D}$ and $\mathit{N}$ are the number of classes and expected maximum number of instances respectively.
We approximate this continuous representation with an NN $\mathfrak{F}_{\Theta} : (\mathbf{p},\mathbf{d}) \rightarrow (\hat{\sigma}, \hat{\mathbf{c}}, \hat{\mathbf{s}}, \hat{\mathbf{k}})$ by optimizing parameters $\Theta$.

As shown in Fig.~\ref{fig:pipeline}, in order to render color images, each point~$\mathbf{p}$ sampled along each pixel ray is first encoded by a color grid into a feature vector $\mathbf{g}_{c}$.
Then these are decoded into density features and the last element is interpreted as the scene volumetric density~$\hat{\sigma}$.
%Since color is a view dependant magnitude, we encode the view direction~$\mathbf{d}$ and concatenate it to the density features before decoding into the final color prediction~$\hat{\mathbf{c}}$
Since color is view dependent, we encode the view direction~$\mathbf{d}$ and concatenate it to the density features before decoding them into the final color prediction~$\hat{\mathbf{c}}$.

On the other hand, panoptic features~$\mathbf{g}_{p}$ are obtained by correcting the appearance features with~$\Delta\mathbf{g}_{p}$.
Semantic~$\hat{\mathbf{s}}$ and instance id~$\hat{\mathbf{k}}$ magnitudes are then directly decoded by shallow NNs as they are view-independent magnitudes. 
Finally, we use the estimated~$\hat{\sigma}$ to perform volumetric rendering of all magnitudes by ray-marching all sampled rays. 

\subsection{Learnable grid encodings}
With the aim of reproducing fine-grain detail while maintaining low running time and memory footprint, we choose to use 3D multi-resolution permutoedral hash-encodings~\cite{rosu2023permutosdf}.
%Since these encoding partitions 3D space into tetrahedral lattices, queried grid features are only interpolated between 4 values (instead of 8 for cubic grids) making them faster than prior approaches.
This encoding partitions 3D space into tetrahedral lattices and the queried grid features are interpolated between 4 values (instead of 8 for cubic grids) making them faster.
%The authors of~\cite{rosu2023permutosdf} provide an open source implementation with custom CUDA kernels.
Furthermore, we book-keep a 3D occtree to further accelerate inference by ray-sampling coordinates in high-density voxels only.
Similar to~\cite{rosu2023permutosdf}, we encode each queried ray sample by interpolating feature vectors at each grid resolution and concatenating them.
Then, we decode them with shallow NNs to obtain the estimated scene properties as shown in Fig.~\ref{fig:pipeline}.

\subsubsection{Delta grid architecture}
As shown by~\cite{panoptic_lifting_siddiqui2023panoptic}, despite the underlying scene geometry being the same, features required for panoptic decoding $(\hat{\mathbf{s}},\hat{\mathbf{k}})$ might be slightly different than the ones needed for appearance decoding $(\hat{\mathbf{c}},\hat{\sigma})$.
Thus, similar to other works~\cite{chen2022tensorf, rosu2023permutosdf}, we choose to have a separate grid encoder for specific outputs, in our case for panoptic quantities.
% However, assuming that both the appearance and panoptic encoders will have many similarities due to the underlying 3D structure, a mechanism to only apply small corrections to the encoding when needed would be optimal.
Our panoptic grid architecture (Fig.~\ref{fig:pipeline}) leverages the similarity between modalities by computing panoptic grid features as~$\mathbf{g}_{p} = \mathbf{g}_{c} + \Delta\mathbf{g}_{p}$.
Where $\Delta\mathbf{g}_{p}$ is a feature vector output of the panoptic grid $G_{p}$.
Thanks to the implicit sparseness of hash-grids, we are able to reduce the capacity $G_{p}$ w.r.t. $G_{c}$, to only have valid values where corrections are needed to have a good panoptic representation.
In addition, we avoid propagating gradients from the panoptic to the color branch to ensure $G_{p}$ only learns corrections on top of $G_{c}$.
See Sec.~\ref{sec:gridAblation} for the corresponding ablations exploring this design choice.

\subsection{Volumetric rendering}
To predict a pixel color from a given camera with center of projection at $\mathbf{o} \in \mathbb{R}^3$ and direction to the pixel $\mathbf{d}\in \mathit{S^{2}}$, NeRF~\cite{nerf_mildenhall2021nerf} leverages volumetric rendering~\cite{volumetric_niemeyer2020differentiable}, integrating the values of field $\mathfrak{F}$ along a ray $\mathbf{r}(t) = \mathbf{o}+t\mathbf{d}$.

\subsubsection{Appearance rendering}
The observed color $C$ depends on viewing direction $\mathbf{d}$, due to illumination and translucence phenomena and so the predicted color $\hat{C}$ can be computed as:
%Given that the observed color $C$ depends on viewing direction $\mathbf{d}$, due to illumination and translucence phenomena, the predicted color $\hat{C}$ can be computed as:
\begin{equation}
    \hat{C}(\mathbf{r},\mathbf{d}) = \int_{t_{i}}^{t_{f}} T(t) \hat{\sigma}(\mathbf{r}(t))\hat{\mathbf{c}}(\mathbf{r}(t),\mathbf{d}))dt \ ,
\label{eq:colorRender}
\end{equation}
\begin{equation}
    T(t) = \textup{exp} \left (\int_{t_{i}}^{t} - \hat{\sigma}(\mathbf{r}(m))dm \right ) \ ,
\end{equation}
where $T(t)$ represents the transmittance probability at $t$.% along the ray.

\subsubsection{Panoptic rendering}
Our model combines semantic and instance predictions at each rendered pixel to produce panoptic predictions.
Unlike color, panoptic fields are independent of view direction, thus semantic and instance predictions can be expressed as continuous functions conditioned only on 3D points $\mathfrak{S}(\mathbf{p})$ and $\mathfrak{I}(\mathbf{p})$.
In order to treat samples along rays as samples of a distribution, similar to \cite{panoptic_lifting_siddiqui2023panoptic}, we apply softmax before ray integration.
Using $\hat{X}$ as a notation proxy for panoptic predicted quantities and $\hat{\mathbf{x}}_{sm} = softmax(\hat{\mathbf{x}})$, $\hat{X}$ can be rendered with the following equation:
\begin{equation}
  \hat{X}(\mathbf{r}) = \int_{t_{i}}^{t_{f}} T(t) \hat{\sigma}(\mathbf{r}(t))\hat{\mathbf{x}}_{sm}(\mathbf{r}(t))dt \ .
\end{equation}

\subsection{Training losses} 
\subsubsection{Color loss}
For $\mathcal{L}_{color}$ we minimize the average 
photometric loss $\| C_{\mathbf{r}} - \hat{C}_{\mathbf{r}} \|^{2}$ over a batch of rays~$\mathbf{r} \in \mathcal{R}$, randomly sampled rays across camera frames.
\subsubsection{Semantic loss}
To predict semantic labels from frame-wise detections of a still-image segmentation model, as proposed by~\cite{semantic_nerf_zhi2021place}, we compute the cross entropy loss between the rendered semantic multi-variate distribution~$\hat{S}_{\mathbf{r}}$ over~$D$ classes and the detected semantic class~${S}_{\mathbf{r}}$ for a batch of rays~$\mathcal{R}$,
%Again for a batch of rays~$\mathcal{R}$, we can express our semantic loss as:
%For a batch of rays~$\mathcal{R}$ the semantic loss is:
 \begin{equation}
      \mathcal{L}_{sem} = \frac{1}{\left| \mathcal{R} \right|} \sum_{\mathbf{r} \in \mathcal{R}} w_{\mathbf{r}} S_{\mathbf{r}} \log \hat{S}_{\mathbf{r}} \ ,
\end{equation}
where $w_{\mathbf{r}}$ is the confidence of the semantic detector~\cite{panoptic_lifting_siddiqui2023panoptic}.
% \subsection{Color loss} 
% For each ray~$\mathbf{r}$ we learn the color predicitons~$\hat{\mathbf{c}_{\mathbf{r}}}(\mathbf{p}_{\mathbf{r}},\mathbf{d}_{\mathbf{r}})$ by minimizing photometric loss between the integrated color~$\hat{C}_{\mathbf{r}}(\mathbf{r}_{\mathbf{r}},\mathbf{d}_{\mathbf{r}})$ and the corresponding RGB pixel value~$C_{\mathbf{r}}$. For a batch~$\mathcal{R}$ of randomly sampled rays across camera frames:
%     \begin{equation}
%          \mathcal{L}_{color} = \frac{1}{\left| \mathcal{R} \right|} \sum_{\mathbf{r} \in \mathcal{R}} \left\| C_{\mathbf{r}} - \hat{C}_{\mathbf{r}} \right\|^{2} \ .
%     \end{equation}
% \subsection{Semantic loss}
% Our model is trained to predict semantic labels from frame-wise detections of still-image segmentation model.
% As proposed by~\cite{semantic_nerf_zhi2021place}, at a ray~$\mathbf{r}$ we compute the cross entropy loss between the rendered semantic multi-variate distribution~$\hat{S}_{\mathbf{r}}$ over~$C$ classes and the detected semantic class~${S}_{\mathbf{r}}$. Again for a batch of rays~$\mathcal{R}$, we can express our semantic loss as:
% \begin{equation}
%      \mathcal{L}_{sem} = \frac{1}{\left| \mathcal{R} \right|} \sum_{\mathbf{r} \in \mathcal{R}} w_{\mathbf{r}} S_{\mathbf{r}} \log \hat{S}_{\mathbf{r}} \ ,
% \end{equation}
% where $w_{\mathbf{r}}$ is the confidence of the semantic detector at ray $\mathbf{r}$ as proposed in~\cite{panoptic_lifting_siddiqui2023panoptic}.

\subsubsection{Instance ID Linear assignment}

To train our model on frame-wise panoptic segmentation predictions, we need to tackle the inter-frame inconsistency of ``things'' IDs. 
Similar to~\cite{panoptic_lifting_siddiqui2023panoptic}, for the $i$-th frame we perform an optimal linear assignment of sampled rays from the~$n$-th predicted thing mask ID~$K^{i}_{n}$ to the~$m$-th most similar rendered one~$\hat{K}^{i}_{m}$.
The assignment cost can be expressed as:
\begin{equation}
     \mathcal{C}^{i}_{nm} = \frac{-1}{\left| \mathcal{R}^{i}_{n} \right|} \sum_{\mathbf{r} \in \mathcal{R}^{i}_{n}} \hat{K}^{i}_{\mathbf{r}m} \ ,
\label{ec:cost}
\end{equation}
%where $\mathcal{R}^{i}_{n}$ is a batch of rays sampled from the~$n$-th detected thing.
We employ the Hungarian algorithm to optimally solve the assignment, obtaining frame-wise pseudo-label vectors~$K'^{i}$. 
This is then used to compute a frame-wise cross entropy loss to train our instance ID head
%At this point we can compute a frame-wise cross entropy loss to train our instance ID head as follows:
%With this we compute a frame-wise cross entropy loss to train our instance ID head
\begin{equation}
     \mathcal{L}_{things} = \frac{1}{\left| \mathcal{I} \right|} \sum_{i \in \mathcal{I}}
     \frac{1}{\left| \mathcal{R}_{i} \right|} \sum_{\mathbf{r} \in \mathcal{R}_{i}}
     w_{\mathbf{r}} K'^{i}_{\mathbf{r}} \log \hat{K}^{i}_{\mathbf{r}} \ ,
\end{equation}
where~$\mathcal{I}$ is a batch of training frames.
We also minimize the following loss for all rays~$\mathcal{R}_{s}$ corresponding to detected stuff classes over all images for their predicted ID to be~$0$.
\begin{equation}
     \mathcal{L}_{stuff} = \frac{1}{\left| \mathcal{R}_{s} \right|} \sum_{\mathbf{r} \in \mathcal{R}_{s}}
     w_{\mathbf{r}} \mathbf{e}_{1} \log \hat{K}^{i}_{\mathbf{r}} \ ,
\end{equation}
where $\mathbf{e}_{1}$ is the first canonical versor.
This yields the following instance ID loss $\mathcal{L}_{id} = \mathcal{L}_{things} + \mathcal{L}_{stuff}$. %Additionally, we apply a segment consistency loss $\mathcal{L}_{K}$ as proposed by~\cite{panoptic_lifting_siddiqui2023panoptic} to instances instead of semantic segments.

\subsubsection{Repeated ID rejection}
\label{sec:idReject}
% \begin{figure}[!t]
%     % \vspace{12pt}
%     \centering
%     \includegraphics[width=0.5\textwidth]{figures/bboxesExample.png}
%     % \vspace{-30pt}
%     \caption{A) Plain linear assignment gives fruits close to the image edges the same ID. B) We solve this by limiting the IDs according to each fruit's position.}
%     \label{fig:frustum_edge}
%     \vspace{-18pt}
% \end{figure}
%Typically in camera setups with planar motion only parallel to the targets (see sec.\ref{sec:datasets}) the scene is measured at a fixed scale.
In our scenario, the camera moves parallel to the targets (see sec.\ref{sec:datasets}) and so the the scene is measured at a fixed scale.
Thus, targets at opposite ends of the camera frustum will appear in several frames on their own, but only in a few together.
This does not encourage the optimal assignment to assign different IDs to them and can lead to multiple objects having the same ID.
%This gives the optimal assignment little encouragement to assign different IDs to them, leading in some cases to multiple objects with the same ID.
This can lead to poor panoptic detection results, see Panoptic Lifting in Fig.~\ref{fig:renderComparison}.

We address this issue with a simple sliding window of assignable IDs, linearly dependent of each target's 3D position along the robot trajectory.
This gets incorporated to the optimal ID association by setting the assignment costs (eq.~\ref{ec:cost}) of predicted IDs outside of the window to a prohibitively high value.
%We design the window such that  for targets at a spatial distance close to the frustum length wont overlapping assignable IDs.
Additionally, we design the window such that targets at a spatial distance close to the frustum length (e.g. at opposite ends of the camera frustrum) won't have overlapping assignable IDs.
Finally, the fruit's position is computed by un-projecting their mask pixels with our model's predicted depth.
% Let~$w$ be the maximum number of expected targets at a given position~$x_{t}$, the available IDs window~$\mathcal{K}_{a}$ can be computed as:
% ~
% \begin{equation}
%      \mathcal{K}_{a}(x_{t}) = \left [ k_{0}(x_{t}), k_{0}(x_{t}) + w \right ] \ ,
% \end{equation}
% \begin{equation}
%     k_{0}(x_{t}) = \frac{w + \Delta k}{l_{frame}} x_{t} = m_{k} x_{t} \ ,
% \end{equation}
% where~$l_{frame}$ is the frustum length in meters at the average depth of all targets, and~$\Delta k$ is an interval safety margin to ensure IDs are not repeated for targets at the extremes of the image.
% Once the window extreme reaches the maximum number of predicted IDs~$N$, we employ modulo arithmetic to restart the window ID values:
% ~
% \begin{equation}
%     k_{0}(x_{t}) = m_{k} (x_{t} \% x_{max}) \ ,
% \end{equation}
% \begin{equation}
%     x_{max} = \frac{N - \Delta k }{m_{k}} \ .
% \end{equation}
% \comment{CM to CS: I didn't really understand the above. Let's talk through it and see if it can be shortened.}

% ~
% Finally,~$x_{t}$ of each pseudolabel~$t$ at each frame is computed by un-projecting all its pixels~$Q_{t}$ to the world frame, using the corresponding model predicted depth~$d_{q}$, and averaging its $x$ coordinate as follows:
% ~
% \begin{equation}
%     x_{t}(Q_{t}) = \frac{1}{\left | Q_{t} \right |} \sum_{\mathbf{q} \in Q{t}} \mathbf{e}_1 P^{-1}\pi^{-1}(\mathbf{q},d_{\mathbf{q}}) \ ,
% \end{equation}
% where $\pi^{-1}$ is the camera inverse projection function which accounts for the camera intrinsics, and $P$ are the camera extrinsics. 

\subsubsection{Post-processing and total loss}
Inspired by~\cite{panoptic_lifting_siddiqui2023panoptic}, we apply a very weak segment consistency loss $\mathcal{L}_{K}$ to instances instead of semantic segments.
%We apply a very weak segment consistency loss $\mathcal{L}_{K}$ as proposed by~\cite{panoptic_lifting_siddiqui2023panoptic} to instances instead of semantic segments.
% Given a set of mask segments $\mathcal{M}$ The regularization loss for a proxy panoptic quantity~$X$ is computes as follows:
%     ~
%     \begin{equation}
%          \mathcal{L}_{rX} = \frac{1}{\left| \mathcal{M} \right|}
%          \sum_{\mathcal{R} \in \mathcal{M}}
%          \mathbf{e}_{n'\mathcal{R}}
%          \sum_{r \in \mathcal{R}} \log \hat{X}_{r}
%     \end{equation}
%     \begin{equation}
%     n'\mathcal{R} = \max_{n}  \sum_{r \in \mathcal{R}} \mathbf{e}_n \hat{X}_{r}
%     \end{equation}
%     ~
A single erosion dilation stage with a 3x3 kernel is applied as post-processing step of the panoptic output.
Finally, our total loss can be written as:
\begin{equation}
        \mathcal{L} =   \lambda_c \mathcal{L}_{color} + 
                        \lambda_s \mathcal{L}_{sem} + 
                        \lambda_{id} \mathcal{L}_{id} + 
                        \lambda_{reg} \mathcal{L}_{K} \ .
\end{equation}
% \comment{CM to CS: the other two losses for regularisation will need ot be added somewhere, these are $\mathcal{L}_{rS}$ $\mathcal{L}_{rK}$.}

\subsection{Camera extrinsics optimization}

As we aim to represent 3D scenes from images with noisy robot odometry poses, these need to be refined to ensure proper multi-view consistency.
Most NeRF approaches perform offline bundle adjustment on the data as a pre-processing step.
This is reasonable when camera extrinisics are unknown.
In our case we have good initial guesses from odometry which allows us to perform online optimization within our training process, similar to~\cite{lin2021barf, kelly2023target}.
To achieve this we add the camera pose parameters to the optimizer and propagate gradients through the color branch and the ray-tracing operation.
Thus, rays~$\mathbf{r}$ in eq.~\ref{eq:colorRender} are made dependent on their corresponding camera transform~$P\!\in$SE3
\begin{equation}
  \mathbf{r}(t,P) = \mathbf{o}_{P}+t R_{P} \mathbf{d}_{c} = \mathbf{o}_{P}+t \mathbf{d} \ ,
\label{eq:pose}
\end{equation}
where $\mathbf{o}_{P}$ and $R_{P}$ are the camera translation and rotation in the world coordinate frame respectively, and $\mathbf{d}_{c}$ is the direction of a pixel in the camera coordinate frame.
% Plugging~\ref{eq:pose} into~\ref{eq:colorRender}, the estimated rendered color can be computed as:
% \begin{equation}
%  \hat{C}(P,\mathbf{d}) = \int_{t_{i}}^{t_{f}} T(t) \hat{\sigma}(\mathbf{r}(t,P))\hat{\mathbf{c}}(\mathbf{r}(t,P),\mathbf{d}))dt \ .
% \label{eq:colorRenderExtrinisics}
% \end{equation}
During optimization we use the 6DOF representation proposed by~\cite{6dofRep_zhou2019continuity} which is suitable for NN optimization.
We do not employ a coarse-to-fine modulation approach~\cite{lin2021barf} as our robot odometry provides a close enough initial estimate of the poses. % to directly optimize camera extrinsics jointly with the 3D representation without additional modifications. 
%In~\cite{lin2021barf}, a coarse-to-fine modulation of the positional encoding was needed to jointly optimize camera extrinsics and a NeRF representation.
%In our case robot odometry provides a close enough initial guess to directly optimize camera extrinsics jointly with the 3D representation without additional modifications.

% \input{contribs/datasets}

\section{Experimental Setup}
\label{sec:datasets}

%\comment{CM to CS: I think all of the subsection headings eats up a lot of space, I'd recommend reducing the number of sub-headings.}

%\subsection{Horticulture Dataset: Sweet Pepper (BUP20)}

We evaluate our models in a very challenging agricultural dataset of glasshouse sweet pepper cropping (BUP20) that we introduced previously in~\cite{smitt2021pathobot}.
It consists of RGB-D images captured with an Intel RealSense D435i sensor by the autonomous phenotyping platform PATHoBot, driving at $0.2 m/s$ in a commercial-like glasshouse environment in Campus Klein Altendorf of the University of Bonn.
% Moreover, the dataset present high detail panoptic segmentation labels.
For this work, we also pre-computed panoptic predictions using an instance segmentation model~\cite{mask2former_cheng2022masked} trained on these domains.
% \begin{figure}[b]
% 	\vspace{-10pt}
% 	\centering
% 	\includegraphics[width=\linewidth]{figures/datasetRobot.png}
% 	\caption{PATHoBot recording data in the field and example color and depth images from the BUP20 dataset, showing the large amount of fruit in the scene under heavy occlusions.}
% 	\label{fig:dbexample}
% 	\vspace{-10pt}
% 	%\vspace{-16pt}
% \end{figure}
% A collection of 10 sequences are included, with two sweet pepper cultivar \textit{Mavera} (yellow) and \textit{Allrounder} (red). Both Cultivars matured from green, to mixed, to their primary color. 
% Sequences are 2 min long sequences of RGB-D images (@15fps, 1280x720), as well as IMU data and wheel odometry.
% In this case we evaluate our models on a panoptic segmentation task with only 2 semantic classes (fruit and background).
The dataset has sparse non-overlapping instance segmentation annotations, and we generate short sequences around the labeled frames for training and validation.
In order to assess panoptic segmentation performance in the plant row closest to the robot we filter out masks with depth larger than 1.5m, ignoring masks from further crop rows, similar to~\cite{smitt2021pathobot}.

\subsection{Implementation details}
PyTorch is used to implement our models and we leverage the Kaolin-Wisp framework~\cite{KaolinWispLibrary} which provides rendering infrastructure, state-of-the-art NeRF building blocks and allows for rapid-ptotyping and integration of new ones.
%We implement our models in Pytorch, leveraging the Kaolin-Wisp framework~\cite{KaolinWispLibrary}, which provides rendering infrastructure, state-of-the-art NeRF building blocks and allows for rapid-ptotyping and integration of new ones.
Both our color and panoptic grids have $21$ LODs and a maximum capacity of $2^{18}$.
We encode ray directions with a regular PE for color prediction.
Our density and color decoders are single layer NNs of width $64$, with ReLU and sigmoid activations respectively.
Our density feature vector has width $16$.
The semantic and instance decoders are shallow and narrow NNs of $2$ and $3$ hidden layers respectively and both of width $64$.
Such small NNs are effective enough as most of the scene representation is already encoded in the feature grids.

\subsubsection{Training scheme}
Since we only have sparse panoptic labeled frames in each video sequence, we train our scene representations in windows of frames around each labeled frame.
Every second frame in the window is used for training, and validation appearance metrics are computed in the remaining frames.
For panoptic metrics, we only compute them for the middle labeled frame.
Finally, we average all the windowed results to obtain our final performance metrics.

We train each scene window for~$800$ epochs, sampling~$4096$ rays per image with~$512$ samples along each for the first~$200$ epochs.
After that we prune the occupancy grid every~$200$ epochs and change from ray-tracing to voxel-tracing, taking~$2$ samples at each of the first few occupied voxels along the rays.
This way we refine the scene geometry close to the surface of objects \cite{nerf_mildenhall2021nerf,rosu2023permutosdf}.
% , while drastically reducing the number of samples requires, thus, speeding up the training and inference process.
For the first~$600$ epochs we only train the color head, later adding the panoptic head for the remaining ones.
We use Adam~\cite{kingma2014adam} as the training optimizer with a momentum of $0.9$ and a fixed global learning rate of $0.01$.
Grid encodings have a learning rate of~$1.0$ in order for them to converge faster.
For all extrinsic parameters we set their learning rate to~$0.0001$.
From an early parameter search, we set the loss weights to~$\lambda_{color}=1$, ~$\lambda_{sem}=0.1$, ~$\lambda_{id}=10$ and ~$\lambda_{reg}=0.1$ for all our experiments.
We train with a batch size of $6$ with images at full resolution ($1280$x$720$).
All models for each validation sample were trained on a single A6000 GPU.
% with 48GB of GDDR6 memory.

\begin{table*}[!b]
\vspace{-10pt}
\caption{Comparison of PAg-NeRF with other relevant panoptic and semantic NeRF methods trained on the BUP20 dataset.}
\vspace{-5pt}
\centering
\begin{tabular}{c|cc|ccc|cc|c}
            & ID Rej. & Uncert. & PSNR [dB] $\uparrow$	    & $PQ$ [\%] $\uparrow$	& IoU [\%] $\uparrow$		& Training [min/seq] $\downarrow$ & \makecell{Inference[s/img]} $\downarrow$	& \#Params. $\downarrow$	\\
\hline
Mask2Former~\cite{mask2former_cheng2022masked}               & - & - & -            	& \textcolor{gray}{71.16}           	 	& \textcolor{gray}{80.52}    	    & -       & -             & -             \\
SemanticNerf~\cite{semantic_nerf_zhi2021place}   	         & \xmark & \xmark & 19.01         	& -              	 	& 77.81  		& 123.2     & 61.3          & 0.63M     \\
PanopticLifting~\cite{panoptic_lifting_siddiqui2023panoptic} & \xmark & \cmark & 21.34         	& 56.65           		& 76.28     		& 36.5      & 8.6          & 7.42M     \\
\hline
		 		                                    & \xmark & \xmark & 23.24         	& 66.83           		& 81.41      	& 26.3     &           &      \\
PAg-NeRF(L)			                                 & \cmark & \xmark & 23.34         	& 68.39           		& 81.45      	& 26.6      & 5.6       & 25.21M     \\ 
	                                                 & \cmark & \cmark & \textbf{23.37}  	& \textbf{70.08}  		& \textbf{82.65}      	& 27.2      &           &      \\

\makecell{PAg-NeRF(S) (sec.~\ref{sec:gridAblation})}	                                         & \cmark & \cmark & 21.37         	& 66.10           		& 79.86      	& \textbf{16.7}      &    \textbf{3.9}       & \textbf{0.62M}     \\

\end{tabular}
\label{tab:results}
%\vspace{-15pt}
\end{table*}

\subsubsection{Validation extrinsics optimization}
Since we jointly optimize camera extrinsics along with the scene representation, validation extrinsics can get slightly miss-aligned to the scene.
Thus, every $10$ epochs we optimize the camera extrinsics of all validation frames while freezing all grid and decoder parameters.
This way the validation extrinsics get registered to the scene representation allowing for a direct comparison.

\subsection{Evaluated models}
In order to evaluate the novel-view rendering quality of the evaluated models, we employ peak signal to noise ratio (PSNR).
The panoptic quality mentric ($PQ$) is used to measure the instance and semantic outputs.
Semantic quality alone is also presented using the intersection over union (IoU).
%We also compute the intersection over union (IoU) for the semantic predictions of all models.

We compare PAg-NeRF with similar state-of-the-art approaches that also employ different radiance fields at their core.
Performance compared to the still-image panoptic segmentation model used to generate the training pseudo-labels is also provided; note that the pseudo-labels do not provide consistent IDs across frames.

\textbf{Mask2Former}~\cite{mask2former_cheng2022masked}: the instance-based semantic segmentation model trained fully-supervised on BUP20.

\textbf{Semantic NeRF}~\cite{semantic_nerf_zhi2021place}: the first approach incorporating a semantic head to an NN based NeRF model.
We train this models using poses obtained through SFM.
Results for this model are presented using only render quality (PSNR) and segmentation (IoU).

\textbf{Panoptic Lifting}~\cite{panoptic_lifting_siddiqui2023panoptic}: is a state-of-the-art panoptic 3D scene representation, leveraging~\cite{chen2022tensorf} as its feature grid.
This models is trained with pre-calculated poses and panoptic predictions from Mask2Former.

\textbf{PAg-NeRF(L)}: our large panoptic 3D scene representation model using the proposed panoptic grid architecture.
All variants are trained with online optimized poses and panoptic predictions from Mask2Former.
Three versions of this model are evaluated: a baseline one, another adding repeated ID rejection (sec.~\ref{sec:idReject}), and a third one using Mask2Former confidence to re-weight all panoptic losses.

\textbf{PAg-NeRF(S)}: A faster version of our model with reduced number of parameters (see sec.~\ref{sec:gridAblation}) that still achieves competitive results.

% The Semantic NeRF and panoptic lifting hyperparameters are kept as reported by the authors and train them for $800$ epochs.
% To reduce their training time, we employ an occupancy octree for ray-tracing, pruned every $200$ epochs.

\begin{figure}[!t]
    \centering
    \includegraphics[width=0.48\textwidth]{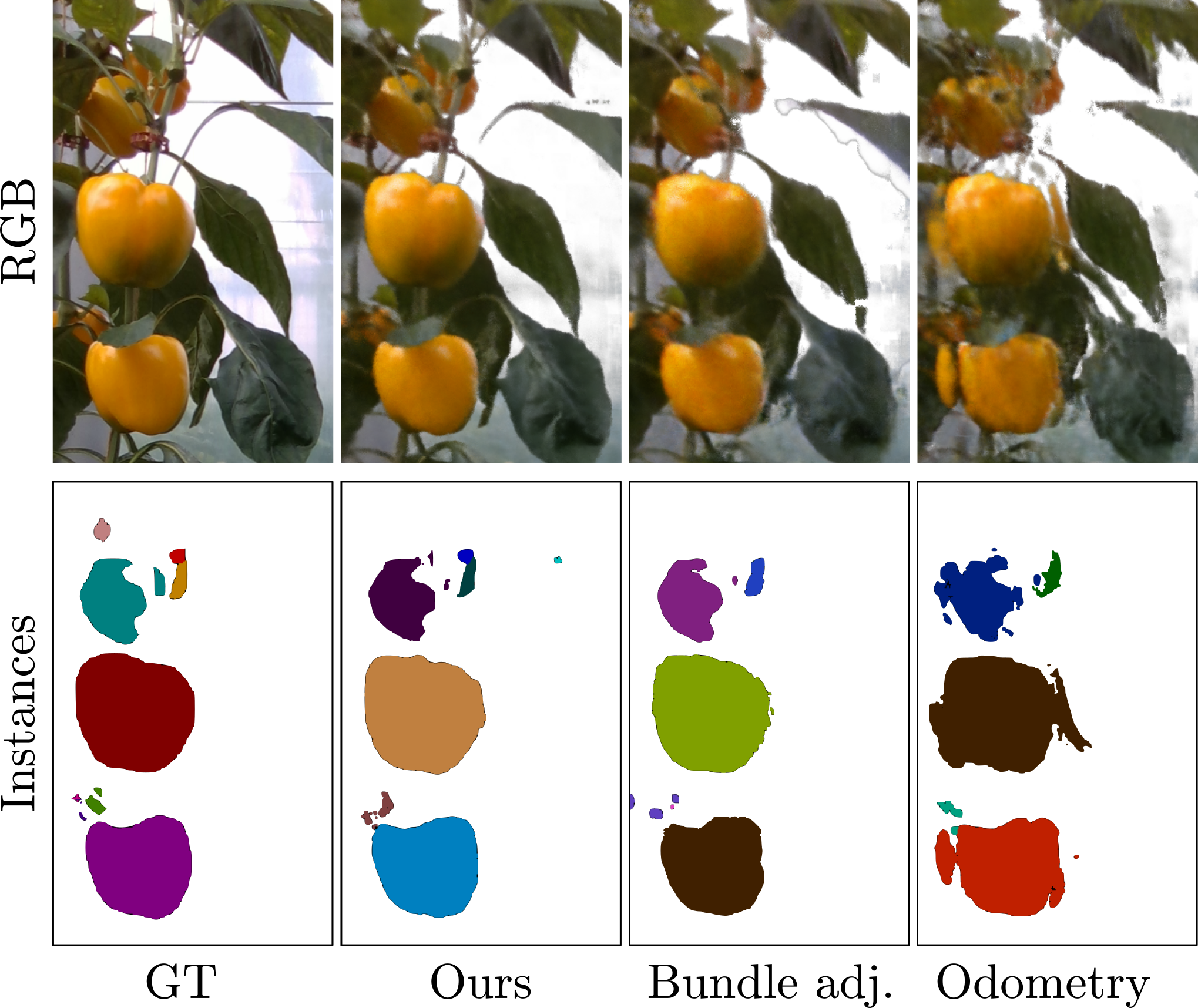}
    \vspace{-5pt}
    \caption{Render comparison for different camera pose sources.}
    \label{fig:pose}
    \vspace{-15pt}

\end{figure}
% \comment{CM to CS: it would be worthwhile to highlight in the ``Instances'' where Ours performs better.
\section{Results}

We start by presenting three quantitative experiments.
First we compare our best model with other semantic novel-view synthesis methods, and the NN still-image detector used for training.
Second, we assess the effect of performing online pose optimization compared to other trajectory sources.
Third, to better understand the benefits and limitations of our panoptic grid architecture we perform an ablation study.
Finally, we showcase qualitative results of novel-view panoptic predictions of our model compared to other systems.
When we present inference time, this is based on how long each model takes to render all outputs per image at full resolution.
%Inference time is based on how long each model takes to render all outputs per image at the dataset's full resolution.

% Moreover, we show our model is capable of consistently maintain instance IDs though multiple views to generate 3D panoptic maps from the implicit representation.

\subsection{Overall performance}

In Tab.~\ref{tab:results} we present the results of our model, Mask2Former, and other relevant NeRF models (SemanticNerf and PanopticLifting).
%Tab.~\ref{tab:results} shows the quantitative comparison of our proposed model with the panoptic detector used for training, as well as other relevant NeRF models with semantic and panoptic outputs.
%It can be seen that our model using uncertainty predictions, the proposed repeated ID rejection loss and pose optimization has the best performance over all, both in terms of $PQ$ and IoU.
It can be seen that the best results are obtained for our model when using uncertainty predictions, the proposed repeated ID rejection loss and pose optimization. % has the best performance over all, both in terms of $PQ$ and IoU.
In particular, our large version of PAg-NeRF outperforms panoptic lifting by 2dB in render quality and an absolute improvement of 13.43 and 6.37 for panoptic and semantics respectively.
Moreover, our system is 1.34 times faster at training and 1.54 times at inference, despite having 3.4 times more parameters than Panoptic Lifting.
This is due to the fast interpolation of permuto grids and shallow NN decoders.
It can also be seen that our modified repeated ID rejections loss 
improves PQ by 1.56 while the use of uncertainty loss re-weighting gives an improvement of another 1.69.

It can also be seen that our large model outperforms semantic NeRF by an absolute IoU margin of 4.84. % 6.35\% in the semantic segmentation tasks.
Furthermore, as our model is grid-based, we are 4.5 and 11 times faster at training and inference respectively.
We attribute this performance gap to our online pose optimization scheme (see sec.~\ref{sec:poseAblation}) and a dedicated feature grid to improve the panoptic representation (see sec.~\ref{sec:gridAblation}).

% An example of the improved detections from incorporating confidence can be seen in fig.~\ref{fig:uncertComparison}, where the system goes from an implicit majority voting scheme to a measurement informed approach, able to disregard false detections, leading to cleaner masks.

It is also worth noting that PAg-NeRF achieves these results with very shallow NN decoders with only a few hidden layers of width $64$, whereas both panoptic lifting and semantic NeRF use more and wider layers of width $256$.

\subsection{Pose estimation ablation}
\label{sec:poseAblation}

Our approach provides an end-to-end system that relies only on initial estimates of pose.
To evaluate the impact of this approach we compare against offline optimized poses with a bundle-adjusted software, and odometry only in terms of render and panoptic quality.
This is because the dataset we use lacks a precise ground-truth trajectory.

%Since the dataset employed lacks a precise trajectory ground-truth, we evaluate the influence of our online pose optimization method over the novel-view synthesis and panoptic segmentation tasks.
%In Tab.~\ref{tab:pose} we show a comparison against offline optimized poses with a bundle-adjusted software, and odometry only.

\begin{table}[!b]
\vspace{-15pt}
\caption{Performance for different camera extrinsic sources.}
\begin{tabular}{llll}
          			       & PSNR{[}dB{]}		& $PQ$ {[}\%{]} & $IoU$ {[}\%{]} \\
\hline
Robot odometry			      & 20.27              & 60.15               			& 79.80         \\
Bundle Adjustment	          & 21.92   		   & 61.54               			& 77.44        \\
Online optimization (Ours)    & \textbf{23.37}     & \textbf{70.08}       			& \ \textbf{82.65}        
\end{tabular}
\label{tab:pose}
% \vspace{-15pt}
\end{table}

In Tab.~\ref{tab:pose} it can be seen that our online pose optimization method produces the best performance.
Compared to bundle-adjusted poses we improve the absolute performance by 1.45dB, 8.54 and 5.21 for render quality, panoptic and semantic metrics respectively.

%w.r.t. , which is one of the most popular approaches to training radiance field-based models.
%Moreover, this can be appreciated in Fig.~\ref{fig:pose}, where our method is able to reproduce fine details of the scene and estimate good instance mask predictions.
Qualitative results in Fig.~\ref{fig:pose} highlight the advantages of our system (Ours) over offline optimized poses with a bundle-adjusted software (Bundle adj.) and odometry only (Odometry).
It can be seen qualitatively that our system (Ours) is able to reproduce fine details of the scene and well estimate instance mask predictions.
In comparison, poses estimated using bundle adjustment (Bundle adj.) can generate good results, however, the resultant renders are noisier with noticeable artifacts for the plant leaves.
%On the other hand, using bundle adjusted poses also generates good results, but produces noisier renders with artifacts that can be seen for the plant leaves.
Finally, it can be seen that using noisy odometry (Odometry) leads to incorrect estimates of the scene geometry presenting smeared panoptic predictions and degraded performance. % \todo{MAH: I don't get this sentence? Where is it shown?}.
% Finally, an example of how noisy odometry can affect prediction quality is shown, where the scene geometry is wrongly estimated, leading to smeared panoptic predictions and degraded performance \todo{MAH: I don't get this sentence? Where is it shown?}.

\subsection{Speed and efficiency ablation study}
\label{sec:gridAblation}
\begin{figure}[!t]
    % \vspace{-12pt}
    \centering
    \includegraphics[width=0.48\textwidth]{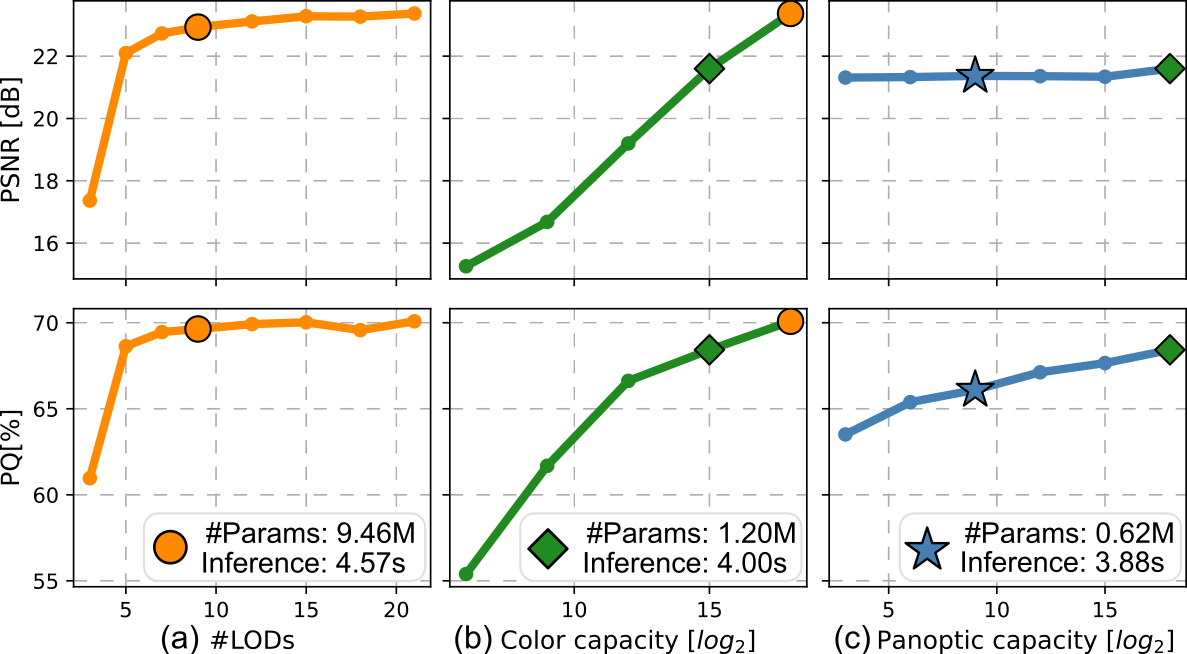}
    \caption{Ablation study of feature grid parameters.}
    \label{fig:gridAblation}
    \vspace{-20pt}
\end{figure}

Seeking to better understand how to tune our model's speed and efficiency, we progressively vary its grid parameters and obtain PAg-NeRF(S), a small yet competitive version of our model.
In Fig.~\ref{fig:gridAblation} we show the render and panoptic performance of our model on 3 successive parameter ablations, where we choose a suitable configuration from one experiment and apply this to the subsequent experiment.
For each experiment, we present the number of parameters and inference time per image to show how we can progressively improve both.   

First, in Fig.~\ref{fig:gridAblation}.a, we take PAg-NeRF(L) (see Tab.~\ref{tab:results}) and reduce the LODs of both grids simultaneously.
We choose to set the number of LODs to $9$ as it still maintains good performance while reducing the number of parameters and inference time by a factor of $2.65$ and $1.2$ respectively. %, when compared to PAg-NeRF(L).

Second, we start from a model with $9$ LODs and progressively reduce the color grid capacity.
As can be seen in Fig.~\ref{fig:gridAblation}.b, this has a large impact on performance and so we choose to keep a high value of ($2^{15}$) for the color capacity.
Despite retaining a relatively high number, we still reduce the number of parameters and improve the inference speed.
%This change further reduces the number of parameters, showing $20M$ less than our largest model and an inference time reduction of $1.4$ times.

Third, the effect of reducing the panoptic delta grid capacity is shown in Fig.~\ref{fig:gridAblation}.c.
As expected, this change has no impact on render quality, since the panoptic branch is detached from the color one.
However, it does have a modest impact on PQ performance.
We chose a panoptic capacity value of $2^{9}$ to further reduce parameter count, trading off some performance.

These optimizations lead to our efficient PAg-NeRF(S) model.
This model has approximately $39$ times fewer parameters than PAg-NeRF(L) and is $1.5$ times faster at inference time.
Furthermore, it still has competitive performance beating the baseline methods (Tab.\ref{tab:results}). 
~

\subsection{Qualitative results}
\begin{figure*}[!th]
    \vspace{12pt}
    \centering
    \includegraphics[width=1\textwidth]{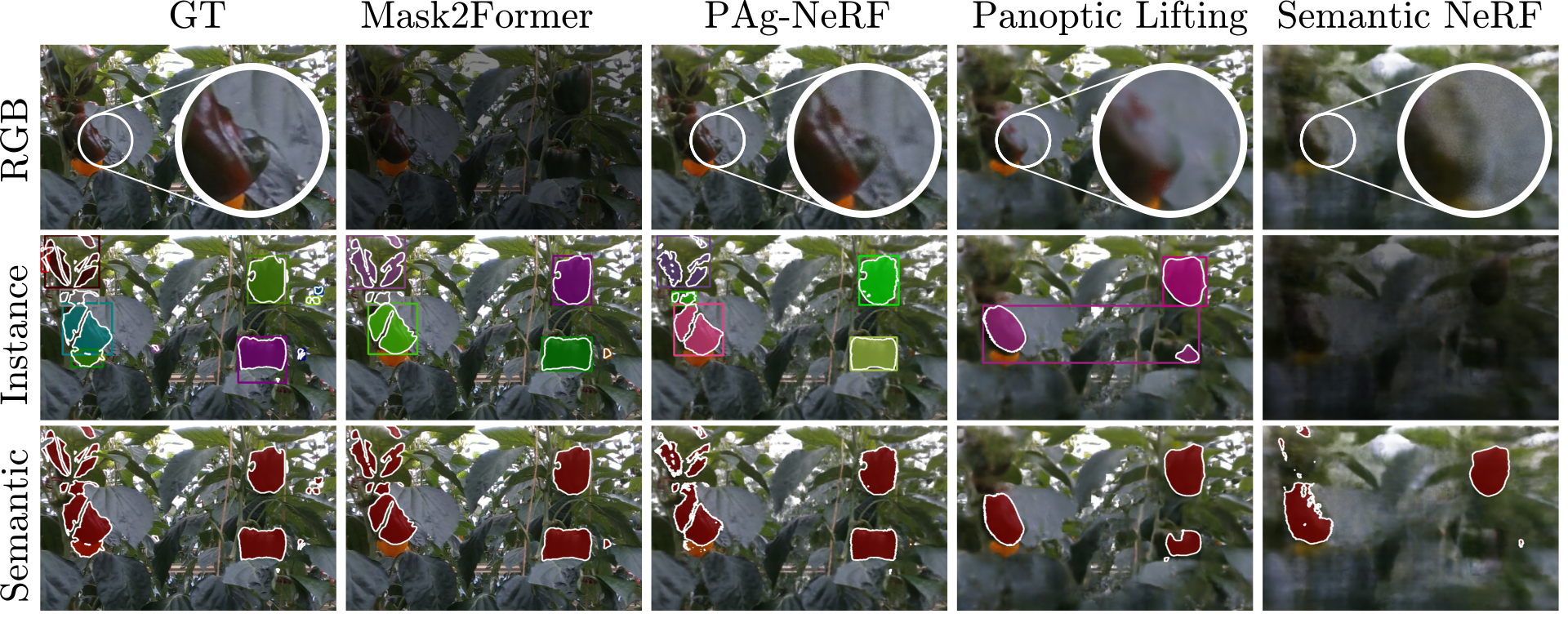}
    \caption{Comparison of PAg-NeRF with the input detections from Mask2Former and other baseline methods. It can be seen how our model is able to reproduce finer details of the scene while producing better panoptic predictions.}
    \label{fig:renderComparison}
    \vspace{-16pt}
\end{figure*}

In Fig.~\ref{fig:renderComparison} we compare the output of PAg-NeRF to Panoptic Lifting, Semantic NeRF and the Mask2Former detector in a very cluttered scene with severely occluded fruits.
In this figure we concentrate on the render (RGB), instance and semantic segmentation quality against the labeled frames.
Further qualitative results for multiple frames and for 3D panoptic maps are available in the supplementary video.
%we provide more qualitative results on video sequences and 3D Panoptic maps extracted from it where PAg-NeRF shines best. 

~
~
In terms of render quality, PAg-NeRF is able to reproduce very fine details of the fruits and leaf textures as well as high frequency edges and thin structures that get smoothed out by Panoptic Lifting.
Moreover, Panoptic Lifting fails to reproduce fine details of masks and misses several fruits.
In the case of Semantic NeRF, it can be seen that some of the fruit and leaves get blended together completely blurring their edges.
% However, some small fruit are missed which is also reflected in our approach's predictions.
% On the other hand, thanks to our repeated ID rejection loss, our model is able to properly distinguish between instances at the far ends of the frame, whereas Panoptic Lifting is not, since it uses a plain linear assignment loss during training.
% Overall, PAg-NeRF is able to achieve accurate scene reconstruction, panoptic segmentation, and sequential id assignment while being a considerably quicker implementation.

The panoptic mask quality (both instance and semantics) for PAg-NeRF is heavily influenced by the detection system, which in this case is Mask2Former.
Generally, Mask2Former produces high quality outputs in this complicated scenario, however, some small fruit are missed which is also reflected in our approach's predictions.
Despite the high quality of Mask2Former, it sometimes misses fruit in a frame before detecting it in a subsequent one, this can be seen in the supplementary video.
In several of these cases, PAg-NeRF is able to recover from these errors through its implicit modelling of scene geometry.
Additionally, thanks to our repeated ID rejection loss, our model is able to properly distinguish between instances at the far ends of the frame, whereas Panoptic Lifting is not as it uses a plain linear assignment loss during training.
Overall, PAg-NeRF is able to achieve accurate scene reconstruction, panoptic segmentation, and sequential ID assignment while being considerably quicker and more memory efficient than Panoptic Lifting.

\section{Summary \& Future Work}

We have presented a novel end-to-end 3D panoptic implicit representation that we validated in a challenging agricultural scenario.
Our architecture is able to distinguish individual fruit instances, being trained only from RGB images with noisy robot poses and still-image panoptic segmentation detections with inconsistent fruit IDs.
By leveraging online pose optimization, our modified instance ID linear assignment loss and hash-permutoedral grid encodings we are able to beat a state-of-the-art 3D panoptic NeRF approach.
Moreover, hash-permutoedral grids allowed us to trade off performance and efficiency, producing fast and efficient models with comparable performance to larger ones.
% Finally we showed panoptic 3D maps that can be extracted from our representation, that could be employed in relevant agricultural tasks such as yield, quality and harvesting time estimation.
As future work, we plan to improve the 3D geometry of our representation and map entire cropping chambers to produce global phenotypic metrics suitable for crop phenotyping.

% \todo{The 3D consistent panoptic representations that is output from this system can be used, in the future, for downstream tasks such as unsupervised/semi-supervised labeling of video streams with consistent identities across frames or for fruit counting (tracking).}

%%%%%%%%%%%%%%%%%%%%%%%%%%%%%%%%%%%%%%%%%%%%%%%%%%%%%%%%%%%%%%%%%%%%%%%%%%%%%%%%
\section*{Acknowledgements}

This work was partially funded by the Deutsche Forschungsgemeinschaft (DFG, German Research Foundation) 459376902 and under Germany’s Excellence Strategy - EXC 2070 – 390732324.

%%%%%%%%%%%%%%%%%%%%%%%%%%%%%%%%%%%%%%%%%%%%%%%%%%%%%%%%%%%%%%%%%%%%%%%%%%%%%%%%%

%\newpage
%\TODO{Search for arxiv references and see if you can find better ones}

\bibliography{references}

\begin{thebibliography}{10}
\providecommand{\url}[1]{#1}
\csname url@rmstyle\endcsname
\providecommand{\newblock}{\relax}
\providecommand{\bibinfo}[2]{#2}
\providecommand\BIBentrySTDinterwordspacing{\spaceskip=0pt\relax}
\providecommand\BIBentryALTinterwordstretchfactor{4}
\providecommand\BIBentryALTinterwordspacing{\spaceskip=\fontdimen2\font plus
\BIBentryALTinterwordstretchfactor\fontdimen3\font minus
  \fontdimen4\font\relax}
\providecommand\BIBforeignlanguage[2]{{%
\expandafter\ifx\csname l@#1\endcsname\relax
\typeout{** WARNING: IEEEtran.bst: No hyphenation pattern has been}%
\typeout{** loaded for the language `#1'. Using the pattern for}%
\typeout{** the default language instead.}%
\else
\language=\csname l@#1\endcsname
\fi
#2}}

\bibitem{ahmadi2021autonomous}
A.~Ahmadi, M.~Halstead, and C.~McCool, ``Towards autonomous visual navigation
  in arable fields,'' in \emph{2022 IEEE/RSJ International Conference on
  Intelligent Robots and Systems (IROS)}.\hskip 1em plus 0.5em minus
  0.4em\relax IEEE, 2022, pp. 6585--6592.

\bibitem{you2022semantics}
A.~You, C.~Grimm, A.~Silwal, and J.~R. Davidson, ``Semantics-guided
  skeletonization of upright fruiting offshoot trees for robotic pruning,''
  \emph{Computers and Electronics in Agriculture}, vol. 192, p. 106622, 2022.

\bibitem{Lehnert20_1}
C.~Lehnert, C.~McCool, I.~Sa, and T.~Perez, ``Performance improvements of a
  sweet pepper harvesting robot in protected cropping environments,''
  \emph{Journal of Field Robotics}, vol.~37, pp. 1197--1223, 2020.

\bibitem{crop_agnostic_halstead2021crop}
M.~Halstead, A.~Ahmadi, C.~Smitt, O.~Schmittmann, and C.~McCool, ``Crop
  agnostic monitoring driven by deep learning,'' \emph{Frontiers in plant
  science}, vol.~12, 2021.

\bibitem{sa2016deepfruits}
I.~Sa, Z.~Ge, F.~Dayoub, B.~Upcroft, T.~Perez, and C.~McCool, ``Deepfruits: A
  fruit detection system using deep neural networks,'' \emph{Sensors}, vol.~16,
  no.~8, p. 1222, 2016.

\bibitem{halstead2018fruit}
M.~Halstead, C.~McCool, S.~Denman, T.~Perez, and C.~Fookes, ``Fruit quantity
  and ripeness estimation using a robotic vision system,'' \emph{IEEE Robotics
  and Automation Letters}, vol.~3, no.~4, pp. 2995--3002, 2018.

\bibitem{zabawa2019detection}
L.~Zabawa, A.~Kicherer, L.~Klingbeil, A.~Milioto, R.~Topfer, H.~Kuhlmann, and
  R.~Roscher, ``Detection of single grapevine berries in images using fully
  convolutional neural networks,'' in \emph{Proceedings of the IEEE Conference
  on Computer Vision and Pattern Recognition Workshops}, 2019, pp. 0--0.

\bibitem{smitt2022explicitly}
C.~Smitt, M.~Halstead, A.~Ahmadi, and C.~McCool, ``Explicitly incorporating
  spatial information to recurrent networks for agriculture,'' \emph{IEEE
  Robotics and Automation Letters}, vol.~7, no.~4, pp. 10\,017--10\,024, 2022.

\bibitem{smitt2021pathobot}
C.~Smitt, M.~Halstead, T.~Zaenker, M.~Bennewitz, and C.~McCool, ``Pathobot: A
  robot for glasshouse crop phenotyping and intervention,'' in \emph{2021 IEEE
  International Conference on Robotics and Automation (ICRA)}.\hskip 1em plus
  0.5em minus 0.4em\relax IEEE, 2021, pp. 2324--2330.

\bibitem{yuepan2023panoptic}
Y.~Pan, F.~Magistri, T.~L{\"a}be, E.~Marks, C.~Smitt, C.~McCool, J.~Behley, and
  C.~Stachniss, ``Panoptic mapping with fruit completion and pose estimation
  for horticultural robots,'' \emph{To be presented at IROS 2023}, 2023.

\bibitem{rosu2023permutosdf}
R.~A. Rosu and S.~Behnke, ``Permutosdf: Fast multi-view reconstruction with
  implicit surfaces using permutohedral lattices,'' in \emph{Proceedings of the
  IEEE/CVF Conference on Computer Vision and Pattern Recognition}, 2023, pp.
  8466--8475.

\bibitem{nerf_mildenhall2021nerf}
B.~Mildenhall, P.~P. Srinivasan, M.~Tancik, J.~T. Barron, R.~Ramamoorthi, and
  R.~Ng, ``Nerf: Representing scenes as neural radiance fields for view
  synthesis,'' \emph{Communications of the ACM}, vol.~65, no.~1, pp. 99--106,
  2021.

\bibitem{semantic_nerf_zhi2021place}
S.~Zhi, T.~Laidlow, S.~Leutenegger, and A.~J. Davison, ``In-place scene
  labelling and understanding with implicit scene representation,'' in
  \emph{Proceedings of the IEEE/CVF International Conference on Computer
  Vision}, 2021, pp. 15\,838--15\,847.

\bibitem{neuralSVF_liu2020neural}
L.~Liu, J.~Gu, K.~Zaw~Lin, T.-S. Chua, and C.~Theobalt, ``Neural sparse voxel
  fields,'' \emph{Advances in Neural Information Processing Systems}, vol.~33,
  pp. 15\,651--15\,663, 2020.

\bibitem{muller2022instant}
T.~M{\"u}ller, A.~Evans, C.~Schied, and A.~Keller, ``Instant neural graphics
  primitives with a multiresolution hash encoding,'' \emph{ACM Transactions on
  Graphics (ToG)}, vol.~41, no.~4, pp. 1--15, 2022.

\bibitem{lin2021barf}
C.-H. Lin, W.-C. Ma, A.~Torralba, and S.~Lucey, ``Barf: Bundle-adjusting neural
  radiance fields,'' in \emph{Proceedings of the IEEE/CVF International
  Conference on Computer Vision}, 2021, pp. 5741--5751.

\bibitem{panoptic_lifting_siddiqui2023panoptic}
Y.~Siddiqui, L.~Porzi, S.~R. Bul{\`o}, N.~M{\"u}ller, M.~Nie{\ss}ner, A.~Dai,
  and P.~Kontschieder, ``Panoptic lifting for 3d scene understanding with
  neural fields,'' in \emph{Proceedings of the IEEE/CVF Conference on Computer
  Vision and Pattern Recognition}, 2023, pp. 9043--9052.

\bibitem{ren2015faster}
S.~Ren, K.~He, R.~Girshick, and J.~Sun, ``Faster r-cnn: Towards real-time
  object detection with region proposal networks,'' in \emph{Advances in neural
  information processing systems}, 2015, pp. 91--99.

\bibitem{panopticSegMatric_kirillov2019}
A.~Kirillov, K.~He, R.~Girshick, C.~Rother, and P.~Doll{\'a}r, ``Panoptic
  segmentation,'' in \emph{Proceedings of the IEEE/CVF conference on computer
  vision and pattern recognition}, 2019, pp. 9404--9413.

\bibitem{panopticFPN_kirillov2019}
A.~Kirillov, R.~Girshick, K.~He, and P.~Doll{\'a}r, ``Panoptic feature pyramid
  networks,'' in \emph{Proceedings of the IEEE/CVF conference on computer
  vision and pattern recognition}, 2019, pp. 6399--6408.

\bibitem{panopticdeeplab_cheng2020}
B.~Cheng, M.~D. Collins, Y.~Zhu, T.~Liu, T.~S. Huang, H.~Adam, and L.-C. Chen,
  ``Panoptic-deeplab: A simple, strong, and fast baseline for bottom-up
  panoptic segmentation,'' in \emph{Proceedings of the IEEE/CVF conference on
  computer vision and pattern recognition}, 2020, pp. 12\,475--12\,485.

\bibitem{transformersurvey_khan2022}
S.~Khan, M.~Naseer, M.~Hayat, S.~W. Zamir, F.~S. Khan, and M.~Shah,
  ``Transformers in vision: A survey,'' \emph{ACM computing surveys (CSUR)},
  vol.~54, no. 10s, pp. 1--41, 2022.

\bibitem{mask2former_cheng2022masked}
B.~Cheng, I.~Misra, A.~G. Schwing, A.~Kirillov, and R.~Girdhar,
  ``Masked-attention mask transformer for universal image segmentation,'' in
  \emph{Proceedings of the IEEE/CVF conference on computer vision and pattern
  recognition}, 2022, pp. 1290--1299.

\bibitem{oneformer_jain2023}
J.~Jain, J.~Li, M.~T. Chiu, A.~Hassani, N.~Orlov, and H.~Shi, ``Oneformer: One
  transformer to rule universal image segmentation,'' in \emph{Proceedings of
  the IEEE/CVF Conference on Computer Vision and Pattern Recognition}, 2023,
  pp. 2989--2998.

\bibitem{sucar2021imap}
E.~Sucar, S.~Liu, J.~Ortiz, and A.~J. Davison, ``imap: Implicit mapping and
  positioning in real-time,'' in \emph{Proceedings of the IEEE/CVF
  International Conference on Computer Vision}, 2021, pp. 6229--6238.

\bibitem{dellaert_kundu2022panoptic}
A.~Kundu, K.~Genova, X.~Yin, A.~Fathi, C.~Pantofaru, L.~J. Guibas,
  A.~Tagliasacchi, F.~Dellaert, and T.~Funkhouser, ``Panoptic neural fields: A
  semantic object-aware neural scene representation,'' in \emph{Proceedings of
  the IEEE/CVF Conference on Computer Vision and Pattern Recognition}, 2022,
  pp. 12\,871--12\,881.

\bibitem{geiger_fu2022panoptic}
X.~Fu, S.~Zhang, T.~Chen, Y.~Lu, L.~Zhu, X.~Zhou, A.~Geiger, and Y.~Liao,
  ``Panoptic nerf: 3d-to-2d label transfer for panoptic urban scene
  segmentation,'' in \emph{2022 International Conference on 3D Vision
  (3DV)}.\hskip 1em plus 0.5em minus 0.4em\relax IEEE, 2022, pp. 1--11.

\bibitem{volumetric_niemeyer2020differentiable}
M.~Niemeyer, L.~Mescheder, M.~Oechsle, and A.~Geiger, ``Differentiable
  volumetric rendering: Learning implicit 3d representations without 3d
  supervision,'' in \emph{Proceedings of the IEEE/CVF Conference on Computer
  Vision and Pattern Recognition}, 2020, pp. 3504--3515.

\bibitem{dvgo_sun2022direct}
C.~Sun, M.~Sun, and H.-T. Chen, ``Direct voxel grid optimization: Super-fast
  convergence for radiance fields reconstruction,'' in \emph{Proceedings of the
  IEEE/CVF Conference on Computer Vision and Pattern Recognition}, 2022, pp.
  5459--5469.

\bibitem{chen2022tensorf}
A.~Chen, Z.~Xu, A.~Geiger, J.~Yu, and H.~Su, ``Tensorf: Tensorial radiance
  fields,'' in \emph{European Conference on Computer Vision}.\hskip 1em plus
  0.5em minus 0.4em\relax Springer, 2022, pp. 333--350.

\bibitem{lod_takikawa2021neural}
T.~Takikawa, J.~Litalien, K.~Yin, K.~Kreis, C.~Loop, D.~Nowrouzezahrai,
  A.~Jacobson, M.~McGuire, and S.~Fidler, ``Neural geometric level of detail:
  Real-time rendering with implicit 3d shapes,'' in \emph{Proceedings of the
  IEEE/CVF Conference on Computer Vision and Pattern Recognition}, 2021, pp.
  11\,358--11\,367.

\bibitem{kelly2023target}
S.~Kelly, A.~Riccardi, E.~Marks, F.~Magistri, T.~Guadagnino, M.~Chli, and
  C.~Stachniss, ``Target-aware implicit mapping for agricultural crop
  inspection,'' in \emph{2023 IEEE International Conference on Robotics and
  Automation (ICRA)}.\hskip 1em plus 0.5em minus 0.4em\relax IEEE, 2023, pp.
  9608--9614.

\bibitem{6dofRep_zhou2019continuity}
Y.~Zhou, C.~Barnes, J.~Lu, J.~Yang, and H.~Li, ``On the continuity of rotation
  representations in neural networks,'' in \emph{Proceedings of the IEEE/CVF
  Conference on Computer Vision and Pattern Recognition}, 2019, pp. 5745--5753.

\bibitem{KaolinWispLibrary}
T.~Takikawa, O.~Perel, C.~F. Tsang, C.~Loop, J.~Litalien, J.~Tremblay,
  S.~Fidler, and M.~Shugrina, ``Kaolin wisp: A pytorch library and engine for
  neural fields research,''
  \url{https://github.com/NVIDIAGameWorks/kaolin-wisp}, 2022.

\bibitem{kingma2014adam}
D.~P. Kingma and J.~Ba, ``Adam: A method for stochastic optimization,''
  \emph{arXiv preprint arXiv:1412.6980}, 2014.

\end{thebibliography}

\bibliographystyle{IEEEtran}

%%%%%%%%%%%%%%%%%%%%%%%%%%%%%%%%%%%%%%%%%%%%%%%%%%%%%%%%%%%%%%%%%%%%%%%%%%%%%%%%%

\end{document}